\documentclass[a4paper,twoside]{article}

\usepackage{epsfig}
\usepackage{subcaption}
\usepackage{calc}
\usepackage{amssymb}
\usepackage{amstext}
\usepackage{amsmath}
\usepackage{amsthm}
\usepackage{multicol}
\usepackage{pslatex}
\usepackage[bottom]{footmisc}

\usepackage{times}
\usepackage{latexsym}
\usepackage{amsmath,amssymb,amsfonts}
\usepackage{algorithmic}
\usepackage{graphicx}
\usepackage{array}
\graphicspath{ {./images/} } 
\usepackage{textcomp}
\usepackage{xcolor}
\usepackage{multirow}
\usepackage{pdfpages}
\usepackage{soul}

\usepackage{tikz}

\usepackage[T1]{fontenc}

\usepackage[utf8]{inputenc}

\usepackage{microtype}

\usepackage[bottom]{footmisc}
\usepackage{natbib}
\usepackage{SCITEPRESS}     

\begin{document}

\title{Can We Use Probing to Better Understand Fine-tuning\\ and Knowledge Distillation of the BERT NLU?}

\author{\authorname{Jakub Ho\'{s}ci\l{}owicz\sup{1,2}\orcidAuthor{0000-0001-8484-1701}, Marcin Sowa\'{n}ski\sup{1,2}\orcidAuthor{0000-0002-9360-1395}, Piotr Czubowski\sup{1}\orcidAuthor{0000-0002-1088-6154} and Artur Janicki\sup{2}\orcidAuthor{0000-0002-9937-4402}}
\affiliation{\sup{1}Samsung R\&D Poland Institute, Warsaw, Poland}
\affiliation{\sup{2}Warsaw University of Technology, Warsaw, Poland}
\email{\{j.hoscilowic, m.sowanski\}@samsung.com, p.czubowski@partner.samsung.com, artur.janicki@pw.edu.pl}
}




\abstract{In this article, we use probing to investigate phenomena that occur during fine-tuning and knowledge distillation of a BERT-based natural language understanding (NLU) model. Our ultimate purpose was to use probing to better understand practical production problems and consequently to build better NLU models. We designed experiments to see how fine-tuning changes the linguistic capabilities of BERT, what the optimal size of the fine-tuning dataset is, and what amount of information is contained in a distilled NLU based on a tiny Transformer. The results of the experiments show that the probing paradigm in its current form is not well suited to answer such questions. Structural, Edge and Conditional probes do not take into account how easy it is to decode probed information. Consequently, we conclude that quantification of information decodability is critical for many practical applications of the probing paradigm. }

\onecolumn \maketitle \normalsize \setcounter{footnote}{0} \vfill

\section{\uppercase{Introduction}}
In recent years significant progress has been made in the natural language processing (NLP) field. Foundation models, such as BERT~\citep{devlin2018bert} and GPT-3~\citep{brown2020language}, remarkably pushed numerous crucial NLP benchmarks and the quality of dialogue systems. Current efforts in the field often assume that progress in foundation models can be made by either increasing the size of a model and dataset or by introducing new training tasks. While we acknowledge that this approach can lead to improvements in many downstream NLP tasks (including natural language understanding -- NLU), we argue that current model evaluation procedures are insufficient to determine whether neural networks ``understand the meaning'' or whether they simply memorize training data. Knowing this would allow us to build better NLU~models.

In the last few decades, measuring the knowledge of neural networks has been one of key challenges for NLP. Perhaps the most straightforward way of doing this is to ask models commonsense questions and measure how often their answer is correct~\citep{bisk2020piqa}. However, as noted in~\citet{bender-koller-2020-climbing}, such an approach might be insufficient to prove whether a neural network ``understands meaning'' because the answers given by foundation models might be the result of memorization of training data and statistical patterns. And since this approach is not conclusive, a new set of methods that focus on explaining the internal workings of neural networks has been proposed~\citep{limisiewicz2020syntax,clark2019does}. Among the most prominent are probing models, which are usually used to estimate the amount of the various types of linguistic information contained in neural network layers~\citep{hewitt2019structural,tenney2019you}. In the probing paradigm, a simple neural network is applied to solve an auxiliary task (e.g., part of speech tagging) to a frozen model. The assumption is that the higher the performance of such a probing neural network, the higher the amount of the respective type of information a probed model contains.

Many efforts have been made to better understand the probing paradigm, its purposes, and its usefulness in generating valuable insights from studies on neural networks. Nevertheless, it is still unclear how exactly probing should be used in the NLP field and what conclusions can be drawn from probing results~\citep{DBLP:journals/corr/abs-2104-08197}. Probing methods are inspired by neuroscience where an analogical approach is used to better understand how information is processed in a human brain~\citep{glaser2020machine}. Most importantly,~\citet{kriegeskorte2019interpreting} conclude that a decoding probe model can only reveal whether a particular type of information is present in a certain brain region. Information perspective is also present in NLP --~\citet{pimentel2020information} state that one should always select the highest-performing probe, because it gives the best estimation of the lower bound of information amount in a neural network.

Our research is motivated by the questions we faced while working on NLU models used in dialogue agents. Development of production-ready BERT-based NLU models requires deep understanding of phenomena that occur during fine-tuning and knowledge distillation (KD). Especially important issues are ``catastrophic forgetting'' and choice of optimal fine-tuning dataset size. Probing seems to be a relevant tool to analyze these issues, but in this work we will show that its practical usability in the NLU context is low. 

Our article is structured as follows. First, in Section~\ref{sec:sota}, we present a review of related work in the area. Next, in Sections~\ref{sec:methods} and~\ref{sec:experiments}, we describe the design and results of our experiments. In Section~\ref{sec:discussion}, we discuss the results of our experiments in the context of related works from both machine learning and neuroscience. Finally, in Section~\ref{sec:conclusions}, we summarize our arguments and outline a proposed solution.

\section{\uppercase{Related Work}}
\label{sec:sota}

In the literature, three types of probing are usually described. \emph{Structural probing}, introduced by~\citet{hewitt2019structural}, which conceptually aims at estimating the amount of syntactic information through reconstruction of dependency trees from vector representations returned by neural networks. \emph{Edge probing}~\citep{tenney2019you} includes a wide array of sub-sentence NLP tasks. All the tasks can be formulated as syntactic or semantic relations between words or sentences. \emph{Conditional probing}~\citep{hewitt2021conditional} is a framework which can be applied to any probing method (e.g., conditional structural probing). Conceptually, conditional probing tells how much information is present in a given layer of a model, provided that this information is not present in a chosen baseline (e.g., embedding layer). Consequently, conditional probing allows for better comparison of models with different architectures -- it grounds probing results in non-contextual embedding layers of models.

Some authors used probing to better understand how linguistic information is acquired by models and how it changes during fine-tuning~\citep{kirkpatrick2017overcoming,hu2020systematic}.~\citet{durrani2021transfer} and~\citet{perez2021much} analyze the phenomenon of “catastrophic forgetting” and generalization in the NLU context.~\citet{zhang2020you} and~\citet{perez2021much} used probing as a tool to estimate the optimal size of pretraining data for NLU tasks.~\citet{zhu2022predicting} concluded that probing is an efficient tool for that, because it does not require gradient updates of the entire model. Probing has also been used to compare amounts of linguistic knowledge in neural networks~\citep{nikoulina2021rediscovery,liu2019linguistic}.

The majority of publications related to BERT fine-tuning report a drop in probing results after fine-tuning.~\citet{durrani2021transfer} conclude that the decrease in probing results means that fine-tuning leads to catastrophic forgetting.~\citet{mosbach2020interplay} hypothesize that after fine-tuning, linguistic information might be less linearly separable and hence harder to detect for a probing model. The broad fine-tuning analysis of~\citet{merchant2020happens} (including probing models) guides authors towards the hypothesis that fine-tuning does not introduce arbitrary (negative) changes to a model's representation, but mainly adjusts it to a downstream task. Relying on improvements and new insights in the probing field (e.g., conditional probing), we wanted to continue along this path in order to understand what happens in our particular NLU scenario.

\begin{figure*} 
    \centering
     \includegraphics[scale=0.39]{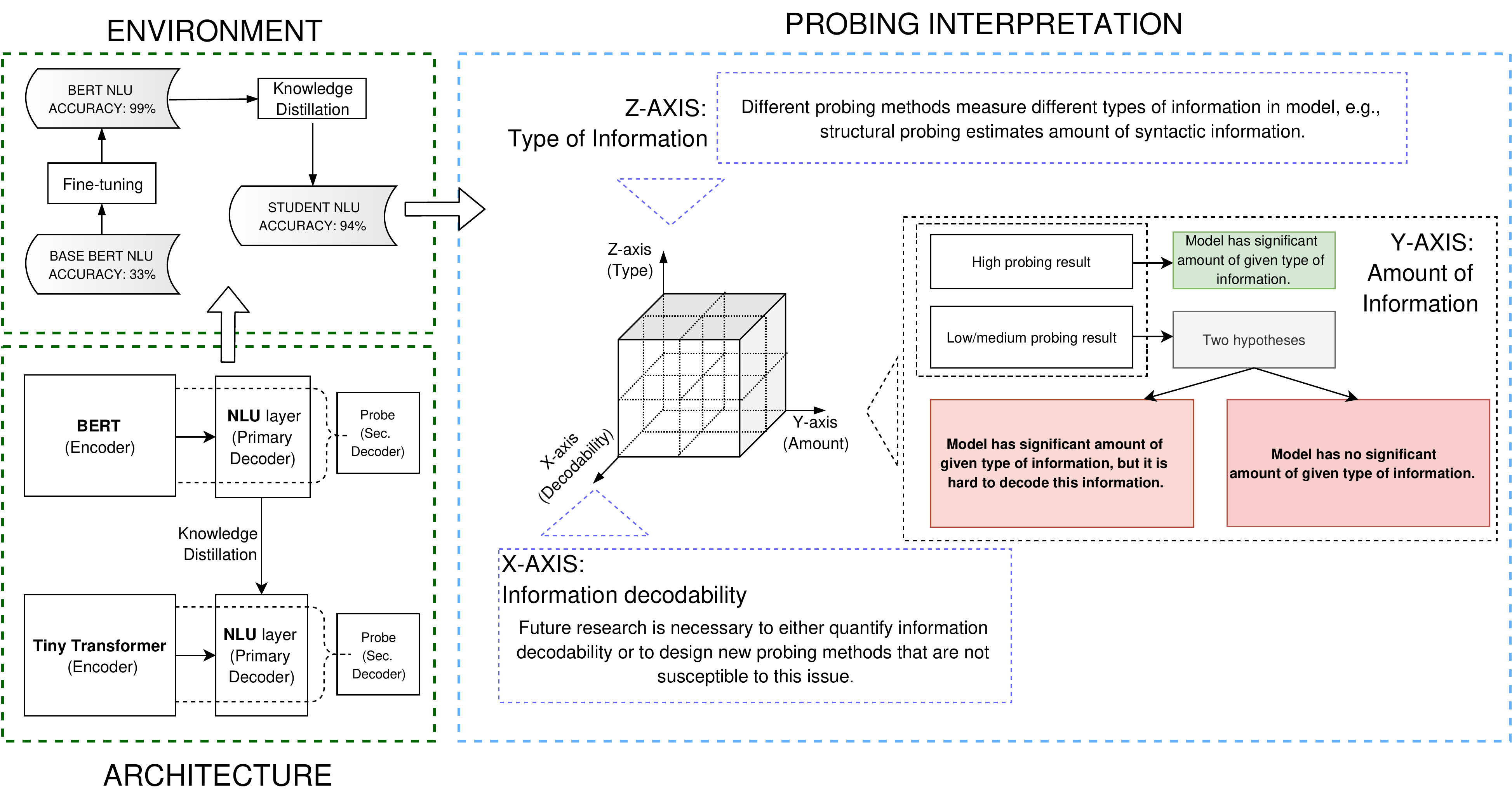}
    \caption{Probing interpretation in typical NLU scenario. The ``Architecture'' sub-graph describes modeling design and terminology. ``Environment'' gives an overview of the pipeline. Finally, the ``Probing Interpretation'' sub-graph describes how probing can be interpreted and what limitations we see.}
    \label{fig:nlu-probing-experiment-diagrams}
\end{figure*}

Probing in the KD context was used by~\citet{kuncoro2019scalable}. The authors concluded that their new neural network architecture has better syntactic competencies than the baseline. Similarly,~\citet{fei2020mimic} used probing to assess the language competencies of a distilled student model. The fact that the probing results of the student model are close to those of the teacher's leads the authors to the conclusion that the student is effective at capturing syntax. In a similar manner, we wanted to use probing to measure the linguistic capabilities of a distilled NLU model.

\section{\uppercase{Methods and Models}}
\label{sec:methods}

To better illustrate our interpretation of probing, we introduced the following terminology (see also Figure~\ref{fig:nlu-probing-experiment-diagrams}): we define the Primary Decoder as the decoder that was used to train or fine-tune a given Encoder (e.g., BERT). For example, in the case of our BERT fine-tuning, it is the NLU head. The Secondary Decoder is an external decoder that was not used for training or fine-tuning of the Encoder (e.g., probing decoder). All probing results noted in this publication are the results of the last layer of the Encoder (e.g., the last layer of BERT).

\subsection{Probing Methods Used}
\label{sec:probing_theory}
We investigated three distinct types of probing methods: structural, edge and conditional.
The performance of the structural probing model is measured with Unlabeled Undirected Attachment Score (UUAS) -- which represents the percentage of undirected edges placed correctly against the gold syntactic tree. We used the part-of-speech (POS) tagging-based variant of edge probing -- its metric is F1 score. As a baseline for conditional probing, we used the non-contextual embedding layer of the probed neural network, similarly to~\citet{hewitt2021conditional}. The metric for conditional probing is determined by the chosen probing method, e.g., for conditional structural probing it is UUAS.

\subsection{Joint NLU} \label{jointnludesc}
There are many NLP tasks that are considered part of the NLU field. In this paper, we define NLU in the way which is most popular in the dialogue agent context -- as the intent classification and slot-filling tasks~\citep{weld2021survey}. An intent represents a command uttered to a dialogue system and slots are parameters of that command. For example, in the utterance ``play Radiohead on Spotify'', `Radiohead' and `Spotify' are slots and `to\_play\_music' is an intent.

As an NLU architecture, we used Joint BERT~\citep{chen2019bert}. This NLU model is created by extending BERT with two softmax classifiers corresponding to intents and slots respectively. To measure NLU quality, we used semantic frame accuracy. It is calculated as a fraction of test sentences where both intent and all slots were correctly predicted. We tested three variants of the NLU architecture, differentiated by the way in which BERT output vectors are passed to the intent classification layer:  

\begin{itemize}
    \itemsep0em
    \item \textbf{Pooled IC}, where input to IC is a vector corresponding to BERT's special [CLS] token. This is the approach used in~\citep{chen2019bert},
    \item \textbf{Average IC}, where input to IC is average of BERT's output vectors,
    \item \textbf{Sum IC}, where input to IC is sum of BERT's output vectors,
\end{itemize}

\subsection{Knowledge Distillation for Joint NLU}

The purpose of KD is to train the student model through imitation of the teacher model. The objective is to minimize KD Loss ${L}_{KD}$, which measures divergence between student, training data and teacher. Because Joint NLU~\citep{chen2019bert} is based on a multitask paradigm, divergence losses of two tasks, intent ${IC}$ classification and slot filling ${SC}$, are minimized simultaneously (${L}_{KD} = {L}_{IC} + {L}_{SC}$). ${L}_{IC}$ is analogical~to:

\begingroup\abovedisplayskip=0.1cm\belowdisplayskip=0.45cm
\begin{align*}
        {L}_{SC}= H(y,\sigma(z_s))+ H(\sigma(z_t; \rho), \sigma(z_s;\rho))
\end{align*}
\endgroup

\noindent{where $H$ is cross entropy loss function, $y$ are ground truth slot labels, $z_s$ and $z_t$ are student and teacher logits respectively. Additionally, softmax function $\sigma$ parametrized by temperature $\rho$, is applied to logits.}

Distillation was performed to a randomly initialized student model with transformer architecture analogical to BERT, but reduced to two layers and two attention heads. BERT NLU (Pooled) was used as the teacher model. As a point of reference, we also included the results of a tiny model with the same architecture as student, but trained without distillation (Tiny Transformer NLU).

\subsection{Data}
We used a subset of Leyzer~\citep{Sowanski2020Leyzer} that consisted of five domains with 51 intents and 29 slots. We annotated Leyzer with the Stanford Dependency Parser~\citep{chen-manning-2014-fast} so that it could be used in the probing context. Additionally, for probing analysis, we used the Universal Dependencies (UD) corpus~\citep{DBLP:journals/corr/abs-2004-10643}, which is manually annotated and consists of general-type sentences unrelated to NLU. The Leyzer subset we used contains 5200 sentences and the UD corpus consists of 16,621 in total. We used an $80\%$, $10\%$, $10\%$ train/test/validation split. To better measure the generalization power of models, we manually constructed a small (164 test cases) Malicious NLU testset. It is based on the Leyzer testset, but consists of sentences which were designed to better measure the generalization power of NLU models. Such test cases contain named entities and grammatical constructs not present in the training set.

\section{\uppercase{Experimental Results}}
\label{sec:experiments}

\subsection{Fine-tuning Analysis}

To get a better perspective on the analyzed phenomena, we evaluated NLU in two variants. In the case of BERT NLU, the BERT model is fine-tuned with NLU head. In the Frozen BERT variant, BERT's weights were fixed during NLU decoder training. Consequently, Frozen BERT gives baseline results (BERT is not fine-tuned). We reported probing results both on Leyzer and UD to give perspective on how in-domain (Leyzer) probing results differ from out-of-domain (UD).

\begin{figure}[h]
        \includegraphics[scale=0.46]{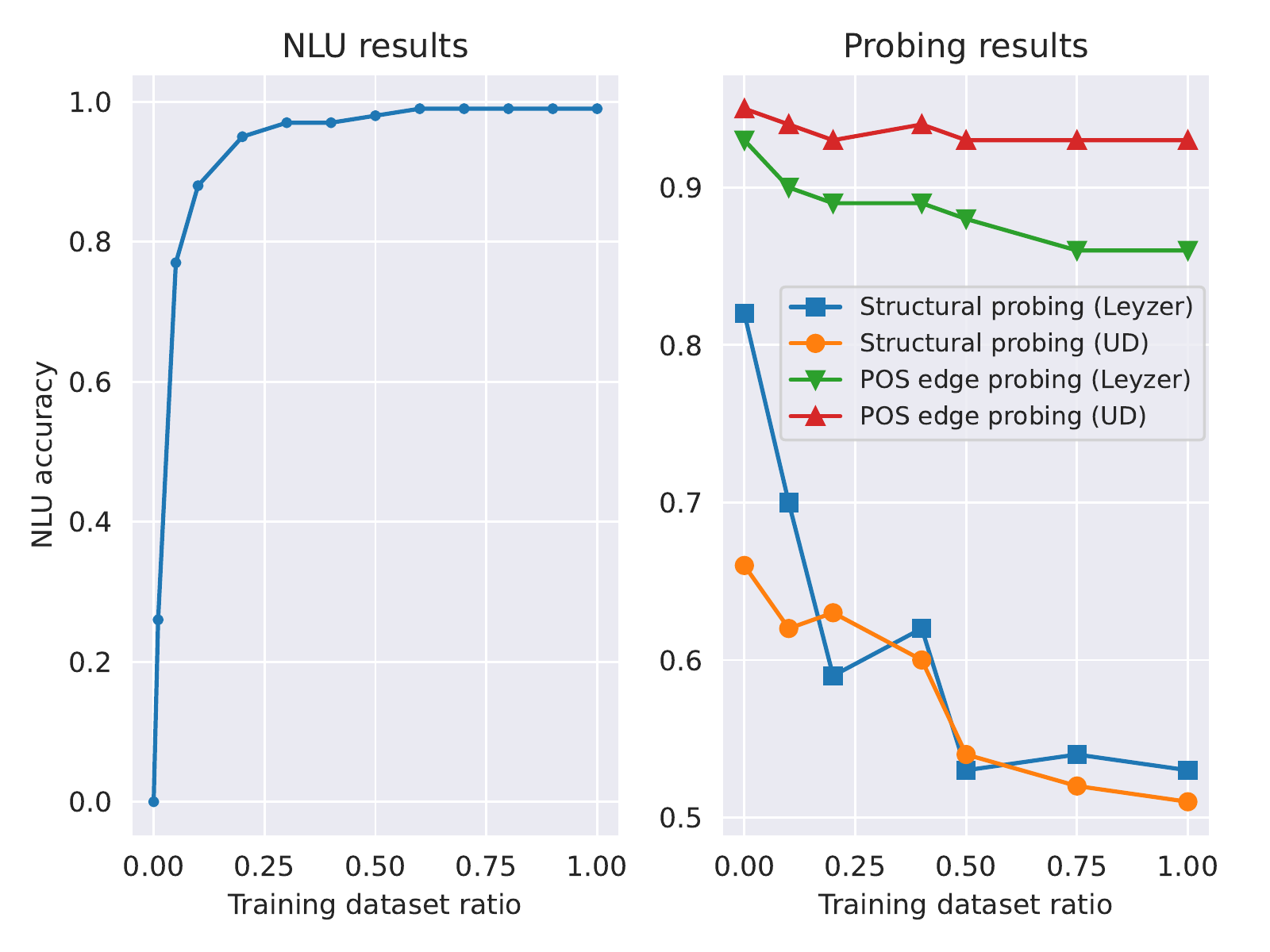}
    \caption{Influence of dataset size on NLU accuracy and probing results.}
    \label{fig:results_dataset_size}
\end{figure}

Detailed results presented in Table~\ref{tbl:nlu-struct_probing-ic_archi_compare} show that, as expected, fine-tuning improved NLU accuracy during the course of training. An inverse trend can be observed in relation to probing results. UUAS gets lower as fine-tuning progresses. In the end, fine-tuned NLU model probing results were significantly lower than those of Frozen BERT. As presented in Figure~\ref{fig:results_dataset_size}, exactly the same tendencies were observed when we gradually increased the size of the fine-tuning dataset.

\begin{table}[h]
\caption{Results of structural probing on BERT NLU model tested on Leyzer and Universal Dependency (UD) testsets. Linear baseline based on \cite{hewitt2019structural}}
\label{tbl:nlu-struct_probing-ic_archi_compare}
\centering
\renewcommand{\arraystretch}{1.2}
\hyphenpenalty=10000
\setlength{\tabcolsep}{12pt}
\scalebox{0.68}{\begin{tabular}{m{3.2cm}m{0.8cm}m{0.8cm}m{0.8cm}m{0.8cm}}
 \hline
  Model Variant & Epoch & NLU Accuracy & Structural probing (Leyzer) & Structural probing (UD)\\
 \hline
 Linear baseline & - & - & - & 0.49 \\ 
 Frozen BERT & 100 & 0.33 & 0.82 & 0.66 \\
 \hline
 \multirow{6}{*}{BERT NLU (Pooled)} & 1 & 0.0 & 0.79 & 0.60 \\
 & 5   & 0.45 & 0.75 & 0.56 \\
 & 10  & 0.85 & 0.70 & 0.55 \\
 & 30  & 0.97 & 0.55 & 0.52 \\
 & 60  & 0.97 & 0.58 & 0.52 \\
 & 100 & 0.97 & 0.50 & 0.50 \\
 \hline
\end{tabular}}
\end{table}

\begin{figure}[t]
    \includegraphics[scale=0.46]{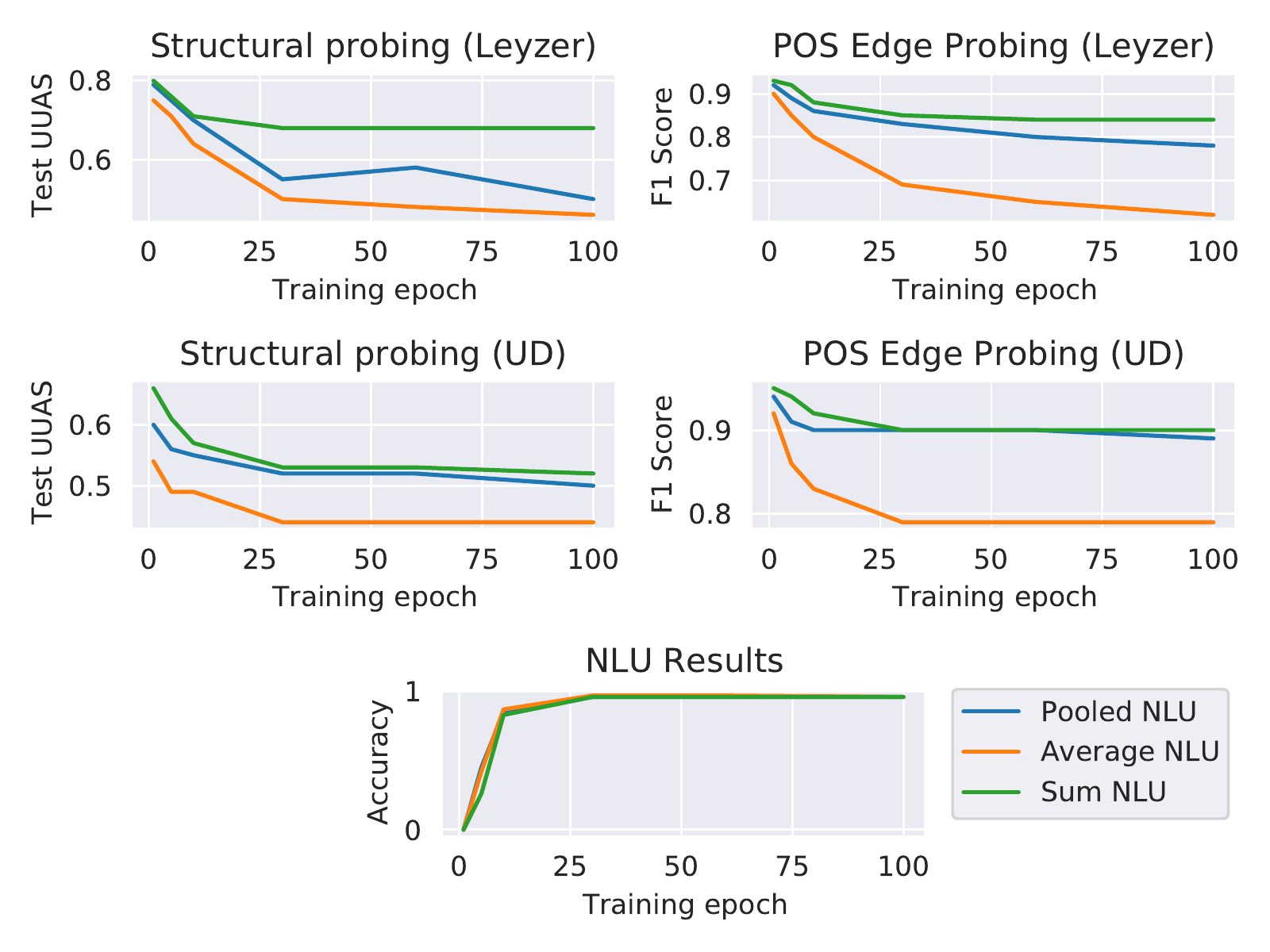}
    \caption{Structural and edge probing results in fine-tuning scenario on Leyzer and UD datasets, for three variants of NLU architecture.}
    \label{fig:fine_tuning-uuas-leyzer-ud}
\end{figure}

In Figure~\ref{fig:fine_tuning-uuas-leyzer-ud}, we present the results of the experiment with three different NLU architectures (as described in subsection~\ref{jointnludesc}). At the end of the fine-tuning process, the NLU results were nearly the same, while the UUAS differed significantly for each architecture. A strong downward trend is visible in all possible cases. All tendencies from this chapter were the same in the case of conditional probing; see example in Table~\ref{tbl:basic_distillation_analysis}.

\subsection{Probing Knowledge Distillation}

The first observation drawn from the NLU results presented in Table~\ref{tbl:basic_distillation_analysis} is that both Student and Tiny Transformer have lower generalization power than BERT NLU. If we compare Student NLU to its non-distilled version, we observe non-negligible test accuracy gain resulting from the KD process. In our case, the temperature was a key factor for distillation -- the best NLU result is achieved for the student distilled with $T=0.01$ (presented in Table~\ref{tbl:basic_distillation_analysis}). Neither temperature nor choice of teacher significantly influences the probing results of the Student NLU.

In Table~\ref{tbl:basic_distillation_analysis} we focused only on conditional probing because it gives a more valuable comparison of models with different architectures; see also Section~\ref{sec:sota}. Both variants of conditional probes indicate that BERT NLU, Student NLU, and Tiny Transformer NLU have a near-zero amount of syntactic and part-of-speech information in the last layers of their Encoders. Specifically, if we measure only information not available in the respective non-contextual baselines, then the amount of information in the last Encoders' layers for all mentioned models is the same and close to zero. We refrain from comparing teacher and student using UD corpus because our student is not pre-trained on any kind of general corpus. Nonetheless, UD results give a scale for conditional probing results. For Frozen BERT, the result on UD corpus is $0.15$ for conditional structural probing and $0.11$ for conditional edge probing.

\begin{table}[h]
\caption{Results for knowledge distillation and probing}
\label{tbl:basic_distillation_analysis}
\centering
\hyphenpenalty=10000
\renewcommand{\arraystretch}{1.2}
\setlength{\tabcolsep}{13pt}
\scalebox{0.65}{
\begin{tabular}{p{3.0cm}p{0.5cm}p{0.9cm}p{0.9cm}p{0.9cm}}
 \hline
    & NLU acc. & Malicious NLU acc. & Conditional Structural Probing (Leyzer) & Conditional Edge Probing (Leyzer)\\
 \hline
 Frozen BERT    & 0.33 & 0.10 & 0.05 & 0.07\\
 \hline
 BERT NLU (Pooled)   & 0.97 & 0.56 & 0.02 & 0.01\\
 Student NLU  & 0.94 & 0.27 & 0.01 & 0.01 \\
 Tiny Transformer NLU    & 0.90 & 0.23 & 0.01 & 0.01 \\
 \hline
\end{tabular}
}
\end{table}

\section{\uppercase{Discussion}}
\label{sec:discussion}

 
The significant decrease in probing results on a general UD corpus can be explained by a known phenomenon: when we fine-tune BERT on an NLU task, its general linguistic knowledge is destroyed (``catastrophic forgetting''). However, the decrease in probing results on the Leyzer corpus is not so easy to interpret. BERT's fine-tuning on Leyzer leads to substantial deterioration of probing results on precisely the same dataset. We initially presumed that fine-tuning on Leyzer would increase related linguistic knowledge, which would be reflected in improvement in the probing results. However, both experiments suggest that fine-tuning optimizes downstream task accuracy, which comes at the expense of the linguistic knowledge associated with Leyzer.

The experiments with different NLU decoder architectures show significant changes in probing results, while at the same time NLU accuracy was not affected that much. Depending on the NLU decoder mode (pooled, averaged, sum), edge and structural probing results can vary from around 0.45 to 0.70 UUAS and $65\%$ to $85\%$ F1 Score. This observation suggests that the NLU decoder can play an important role in how linguistic information in the Encoder is structured.

In the conditional framework, the downward probing tendencies of BERT's fine-tuning do not change. Conditional results on Leyzer are nearly the same for teacher and non-pretrained student NLUs. Both models achieve near-zero conditional probing results and if we apply a standard interpretation, we can conclude that neither contains a significant amount of syntactic and part-of-speech information in the last layers of their Encoders (compared to the respective non-contextual baselines).

All mentioned observations guided us toward the two main hypotheses presented in ``Amount of Information'' in Figure~\ref{fig:nlu-probing-experiment-diagrams}. The deterioration of probing results might mean either that the amount of linguistic information decreased or that it is harder to decode for a probing model.

The purpose of the Primary NLU Decoder is to structure BERT's knowledge so that it can effectively extract it and achieve high accuracy on a downstream task. The Primary Decoder does not know about the existence of Secondary Decoders and its goal is not to make information in the Encoder easily extractable for them, but to reduce NLU training loss. Consequently, making definitive conclusions (``fine-tuning leads to catastrophic forgetting'') relying on probing results could be misleading. The degree of ``catastrophic forgetting'' (as measured with probing) could be much smaller than we think, or even non-existent. However, as shown in Figure~\ref{fig:nlu-probing-experiment-diagrams}, without inclusion of the decodability issue in the probing paradigm, such considerations remain inconclusive. Low probing results can mean either that information is not present, or that it is hard to decode.



\section{\uppercase{Conclusions and Future Work}}
\label{sec:conclusions}

Relying on insights from the experiments, neuroscience~\citep{kriegeskorte2019interpreting,DBLP:journals/corr/abs-2104-08197} and NLP, we conclude that probing has small usability for analysis of NLU models. Current probing methods do not consider how easy it is to decode probed information, hence they only give an estimation of the lower bound of information amount~\citep{pimentel2020information}. Future research is necessary to either quantify information decodability (as visualized in Figure~\ref{fig:nlu-probing-experiment-diagrams}) or to design new probing methods that are not susceptible to this issue.

However, to measure decodability, firstly, information in neural networks must be defined in a rigorous way. Ultimately, similar to~\citep{tschannen2019mutual}, we conclude that a new concept of information is important for the future of NLP research. One approach to how information can be defined in neural networks is presented by~\citet{xu2020theory}, but we have not yet found any works about information decodability.

Another path is to use probing in a neuroscience manner, as proposed by~\citet{kriegeskorte2019interpreting}. Instead of trying to answer the question ``How much information of a given type is in the neural network?'', we can focus on the question ``Is a given type of information present in the neural network?'' Consequently, we use probing results as a component of $p$-value for the hypothesis that there is no significant amount of a given type of information in a given neural network layer. This path implies that we should focus more on new types of information which are less obvious than those currently probed. Information decodability also constitutes an issue for this approach. However, in our opinion, with proper design of hypothesis testing (including heuristics about decodability of information), this approach can give more reasonable insights than the ``How much information'' paradigm.

To summarize, the main contributions of our work are as follows:
\begin{itemize}
     \item{We showed that current probing methods are of low usability for analysis of NLU models. Without careful interpretation, they might lead to wrong conclusions.} 
     \item{We presented a clear interpretation of probing in the NLP context. This interpretation implies that information decodability is a large obstacle for many practical applications of probing methods.}
\end{itemize}

\bibliographystyle{apalike}
\bibliography{bibliography}

\end{document}